\documentclass[11pt,a4paper]{article}
\usepackage[hyperref]{acl2020}
\usepackage{times}
\usepackage{latexsym}

\usepackage{microtype}

\newcommand\new[1]{{#1}}
\newcommand\vinnew[1]{{#1}}

\usepackage{balance}       
\usepackage{graphics}      
\usepackage[T1]{fontenc}   
\usepackage{txfonts}
\usepackage{mathptmx}
\usepackage{color}
\usepackage{booktabs}
\usepackage{dblfloatfix}
\usepackage{float}

\usepackage{textcomp}

\usepackage{caption}

\usepackage{subcaption}

\usepackage{enumitem}

\usepackage{url}
\usepackage{booktabs}
\usepackage{times}
\usepackage{latexsym}

\usepackage{comment}

\usepackage{url}

\usepackage{microtype}        
\usepackage{ccicons}          

\usepackage{todonotes}

\aclfinalcopy 
\def\aclpaperid{2943} 

\title{Social Biases in NLP Models as Barriers 
for Persons with Disabilities}

\author{
Ben Hutchinson,
Vinodkumar Prabhakaran,
Emily Denton,\\
\textbf{Kellie Webster,
Yu Zhong,
Stephen Denuyl}
\\
Google \\
\texttt{\{benhutch,vinodkpg,dentone,websterk,yuzhong,sdenuyl\}@google.com}
}

\date{}

\begin{document}
\maketitle
\begin{abstract}
Building equitable and inclusive NLP technologies demands consideration of whether and how social attitudes are represented in ML models. 
In particular, representations encoded in models often inadvertently perpetuate undesirable social biases from the data on which they are trained.
In this paper, we present evidence of such undesirable biases towards mentions of disability in two different English language models: toxicity prediction and sentiment analysis.
Next, we demonstrate that the neural embeddings that are the critical first step in most NLP pipelines similarly contain undesirable biases towards mentions of disability. 
We end by highlighting topical biases in the discourse about disability which may contribute to the observed model biases;
for instance, gun violence, homelessness, and drug addiction are over-represented in texts discussing mental illness. 
\end{abstract}

\section{Introduction}

This paper focuses on the representation of persons with disabilities through the lens of technology.
Specifically, we examine how NLP models classify or predict text relating to persons with disabilities (see Table~\ref{table_problem}). 
This is important because NLP models are increasingly being used for tasks such as fighting online abuse \cite{perspectiveApi}, \new{measuring brand sentiment \cite{mostafa2013more},} and matching job applicants to job opportunities \cite{de2019bias}.
In addition, since text classifiers are trained on large datasets, the biases they exhibit may be indicative of societal perceptions of persons with disabilities \cite{Caliskan2017SemanticsDA}.
\new{If models inappropriately condition on mentions of disability, this could impact people writing, reading, or seeking information about a disability. 
Depending on how such models are deployed, this could potentially result in reduced autonomy, reduced freedom of speech, perpetuation of societal stereotypes or inequities, or harms to the dignity of individuals.}

While previous studies have studied unintended biases in NLP models against other historically marginalized groups \cite{Bolukbasi2016,Caliskan2017SemanticsDA,Garg2017, barocas2017problem,garg2018counterfactual, dixon2018measuring,noble2018algorithms,manzini2019black,sap-etal-2019-risk,may2019measuring,speer2017}, bias with respect to different disability groups has been relatively under-explored. 
However, over one billion individuals (about 15\% of the world's population) are persons with disabilities,\footnote{https://www.worldbank.org/en/topic/disability} and disability is sometimes the subject of strong negative social biases. 
For example, a 2007 study found implicit and explicit preferences against people with disabilities compared to people without disabilities across the social group domains \cite{nosek2007pervasiveness}.

\begin{table}[t]
\centering
\small
\begin{tabular}{@{}l
cc@{}}
\toprule
Sentence & Toxicity \\
\midrule
I am a person with mental illness. & 0.62 \\
I am a deaf person. & 0.44 \\
I am a blind person. & 0.39 \\
I am a tall person. & 0.03 \\
I am a person. & 0.08 \\
\midrule
I will fight for people with mental illnesses. & 0.54 \\
I will fight for people who are deaf. & 0.42 \\
I will fight for people who are blind. & 0.29 \\
I will fight for people. & 0.14 \\
\bottomrule
\end{tabular}
\caption{Example toxicity scores from Perspective API.
}
\label{table_problem}
\end{table}

\vinnew{In this paper, we study how social biases about persons with disabilities can be perpetuated by NLP models.
} 
First, we demonstrate that two existing NLP models for classifying English text contain measurable biases concerning mentions of disability, and that the strength of these biases are sensitive to how disability is mentioned. 
Second, we show that language models that feed NLP systems for downstream application similarly contain measurable biases around disability.
Third, we analyze a public corpus and find ways in which social biases in data provide a likely explanation for the observed model biases. 
\new{We conclude by discussing the need for the field to consider socio-technical factors to understand the implications of findings of model bias.}

\section{Linguistic Phrases for Disabilities}
\label{sec_phrases}

Our analyses in this paper use a set of 56 linguistic expressions (in English) for referring to people with various types of disabilities,
e.g.\ {\em a deaf person}. We partition these expressions as either \textit{Recommended} or \textit{Non-Recommended}, according to their prescriptive status, by consulting guidelines published by three US-based organizations: Anti-Defamation League, {\sc acm sigaccess} and the ADA National Network \cite{cavender:writing, Hanson2015, adl, ada}. 
\new{We acknowledge that the binary distinction between recommended and non-recommended is only the coarsest-grained view of complex and multi-dimensional social norms, however more input from impacted communities is required before attempting more sophisticated distinctions \cite{jurgens2019}.}
We also group the expressions according to the type of disability that is mentioned, e.g. the category {\sc hearing} includes phrases such as "a deaf person" and "a person who is deaf".
Table~\ref{tab:phrases} shows a few example terms we use. The full lists of 
recommended and non-recommended 
terms are in Tables~\ref{tab:terms_list} and \ref{tab:bad_terms_list} in the appendix.

\section{Biases in Text Classification Models}
\label{sec_psa}

Following \cite{garg2018counterfactual, vinod2019}, we use the notion of {\em perturbation}, whereby the phrases for referring to people with disabilities, described above, are all inserted into the same slots in sentence templates. 
We start by first retrieving a set of naturally-occurring sentences that contain the pronouns {\em he} or {\em she}.\footnote{Future work will see how to include non-binary pronouns.} We then select a pronoun in each sentence, and ``perturb'' the sentence by replacing this pronoun with the phrases described above.  
Subtracting the NLP model score for the original sentence from that of the perturbed sentence gives the \textit{score diff}, a measure of how changing from a pronoun to a phrase mentioning disability affects the model score.

\begin{figure*}
     \centering
     \begin{subfigure}[b]{0.495\textwidth}
        \centering
        \includegraphics[width=\textwidth]{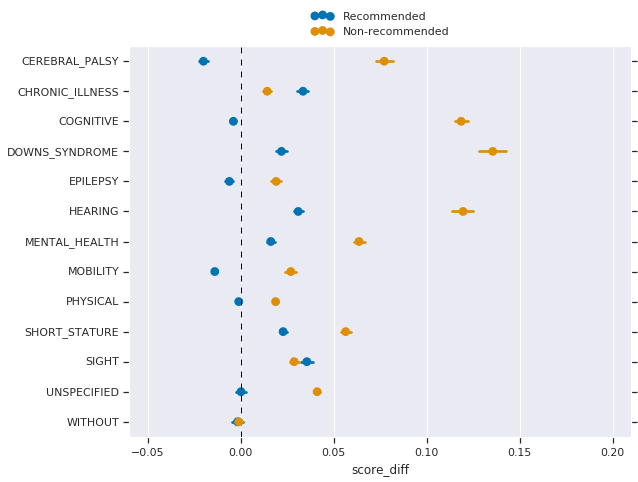}
        \caption{Toxicity model: higher means more likely to be toxic.}
        \label{fig:toxicity_score_diff}
     \end{subfigure}
     \begin{subfigure}[b]{0.495\textwidth}
        \centering
        \includegraphics[width=\textwidth]{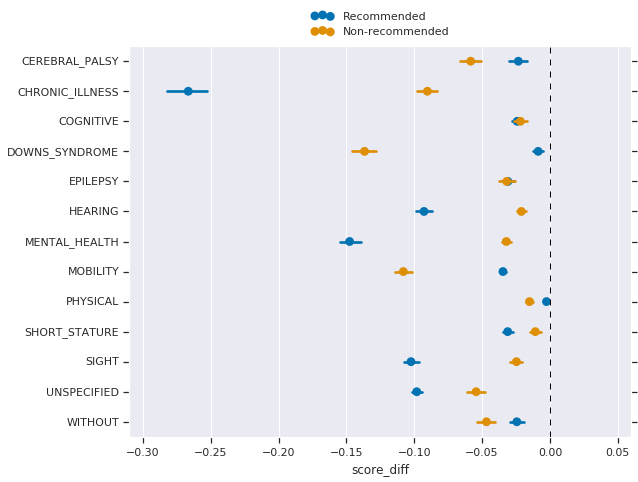}
        \caption{Sentiment model: lower means more negative.}
    \label{fig:sentiment_score_diff}
     \end{subfigure}
    \caption{Average change in model score when substituting a {\em recommended} (blue) or a
    {\em non-recommended} (yellow) phrase for a person with a disability, compared to a pronoun.
    Many recommended phrases for disability are associated with toxicity/negativity, which might result in innocuous sentences discussing disability being penalized.}
    \label{fig:model_score_diff}
\end{figure*}

We perform this method on a set of  1000 sentences extracted at random from the Reddit sub-corpus of \cite{voigt-etal-2018-rtgender}.
Figure~\ref{fig:toxicity_score_diff} shows the results for toxicity prediction \cite{perspectiveApi}, which outputs a score $\in [0,1]$, with higher scores indicating more toxicity. 
For each category, we show the average \textit{score diff} for recommended phrases vs. non-recommended phrases along with the associated error bars. 
All categories of disability are associated with varying degrees of toxicity, while the aggregate average \textit{score diff} for recommended phrases was smaller (0.007) than that for non-recommended phrases (0.057). Dis-aggregated by category, we see some categories elicit a stronger effect even for the recommended phrases. 
Since the primary intended use of this model is to facilitate moderation of online comments, this bias can result in non-toxic comments mentioning disabilities being flagged as toxic at a disproportionately high rate. 
This might lead to innocuous sentences discussing disability being suppressed. 
Figure~\ref{fig:sentiment_score_diff} shows the results for a sentiment analysis model \cite{GoogleCloudNLPAPI} that outputs scores $\in [-1,+1]$; higher score means positive sentiment. Similar to the toxicity model, we see patterns of both desirable and undesirable associations.

\begin{table}[t]
    \centering
    \small

    \begin{tabular}{cl}
    \toprule
    Category                &Phrase \\ \midrule
         \sc sight&a blind person (R)  \\  
         \sc sight&a sight-deficient person (NR)  \\  
         \sc mental\_health & a person with depression (R) \\
         \sc mental\_health & an insane person (NR) \\
         \sc cognitive & a person with dyslexia (R)  \\
         \sc cognitive & a slow learner (NR)  \\
         \bottomrule
    \end{tabular}
    \caption{Example phrases recommended (R) and non-recommended (NR) to refer to people with disabilities.}
    \label{tab:phrases}
\end{table}

\vspace{-0.2mm}
\section{Biases in Language Representations}
\label{sec_bert}

Neural text embedding models \cite{Mikolov2013EfficientEO} are critical first steps in today's NLP pipelines. 
These models learn vector representations of words, phrases, or sentences, such that semantic relationships between words are encoded in the geometric relationship between vectors.
Text embedding models capture some of the complexities and nuances of human language. 
However, these models may also encode undesirable correlations in the data that reflect harmful social biases \cite{Bolukbasi2016, may2019measuring, Garg2017}. 
Previous studies have predominantly focused on biases related to race and gender, with the exception of \citet{Caliskan2017SemanticsDA}, who considered physical and mental illness. Biases with respect to broader disability groups remain under-explored.
In this section, we analyze how the widely used bidirectional Transformer (BERT) \cite{Devlin2018BERTPO}\footnote{We use the 1024-dimensional `large' uncased version, available at \url{https://github.com/google-research/}.} model represents phrases mentioning persons with disabilities. 

Following prior work \cite{kurita-etal-2019-measuring} studying social biases in BERT, we adopt a template-based fill-in-the-blank analysis. 
Given a query sentence with a missing word, BERT predicts a ranked list of words to fill in the blank. We construct a set of simple hand-crafted templates \textit{`<phrase> is  \underline{\hspace{3mm}}.'}, where \textit{<phrase>} is perturbed with the set of \textit{recommended} disability phrases described above.
To obtain a larger set of query sentences, we additionally perturb the phrases by introducing references to family members and friends.
For example, in addition to `a person', we include `my sibling', `my parent', `my friend', etc. 
We then study how the top ranked\footnote{we consider the top 10 BERT word predictions.} words predicted by BERT change when different disability phrases are used in the query sentence.

\begin{figure}
    \centering
    \includegraphics[width=\linewidth]{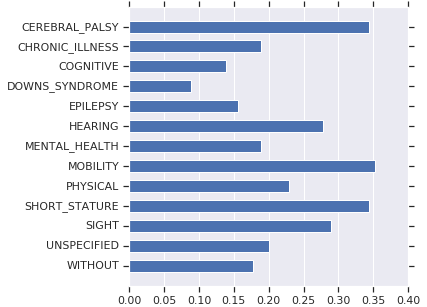}
    \caption{Frequency with which word suggestions from BERT produce negative sentiment score.}
    \label{fig:bert}
\end{figure}

\begin{table*}[]
\centering
\small
\begin{tabular}{@{}lrlrlrlrlr@{}}
\toprule
\sc condition & Score & \sc treatment & Score & \sc infra. & Score & \sc linguistic & Score & \sc social & Score \\ \midrule
mentally ill & 23.1 & help & 9.7 & hospital & 6.3 & people & 9.0 & homeless & 12.2 \\
mental illness & 22.1 & treatment & 9.6 & services & 5.3 & person & 7.5 & guns & 8.4 \\
mental health & 21.8 & care & 7.6 & facility & 5.1 & or & 7.1 & gun & 7.9 \\
mental & 18.7 & medication & 6.2 & hospitals & 4.1 & a & 6.2 & drugs & 6.2 \\
issues & 11.3 & diagnosis & 4.7 & professionals & 4.0 & with & 6.1 & homelessness & 5.5 \\
mentally & 10.4 & therapy & 4.2 & shelter & 3.8 & patients & 5.8 & drug & 5.1 \\
mental disorder & 9.9 & treated & 4.2 & facilities & 3.4 & people who & 5.6 & alcohol & 5.0 \\
disorder & 9.0 & counseling & 3.9 & institutions & 3.4 & individuals & 5.2 & police & 4.8 \\
illness & 8.7 & meds & 3.8 & programs & 3.1 & often & 4.8 & addicts & 4.7 \\
problems & 8.0 & medications & 3.8 & ward & 3.0 & many & 4.5 & firearms & 4.7 \\
\bottomrule
\end{tabular}
\caption{Terms that are over-represented in comments with mentions of the \textit{psychiatric\_or\_mental\_illness} based on the \cite{jigsawkaggle} dataset, grouped across the five categories described in Section~\ref{sec_kaggle}. Score represents the log-odds ratio as calculated using \cite{monroe2008fightin}; a score greater than 1.96 is considered statistically significant.}
\label{tab_kaggle_terms}
\end{table*}

In order to assess the valency differences of the resulting set of completed sentences for each phrase, we use the Google Cloud sentiment model \cite{GoogleCloudNLPAPI}.
For each BERT-predicted word \textit{w}, we obtain the sentiment for the sentence \textit{`A person is <w>'}. We use the neutral \textit{a person} instead of the original phrase, so that we are assessing only the differences in sentiment scores for the words predicted by BERT and not the biases associated with disability phrases themselves in the sentiment model (demonstrated in Section~\ref{sec_psa}). 
Figure~\ref{fig:bert} plots the frequency with which the fill-in-the-blank results produce negative sentiment scores for query sentences constructed from phrases referring to persons with different types of disabilities.
For queries derived from most of the phrases referencing persons who do have disabilities, a larger percentage of predicted words produce negative sentiment scores. 
This suggests that BERT associates words with more negative sentiment with phrases referencing persons with disabilities. 
Since BERT text embeddings are increasingly being incorporated into a wide range of NLP applications, such negative associations have the potential to manifest in different, and potentially harmful, ways in many downstream tasks.

\section{Biases in Data}
\label{sec_kaggle}

NLP models such as the ones discussed above are trained on large textual corpora, which are analyzed to build ``meaning'' representations for words based on word co-occurrence metrics, drawing on the idea that ``you shall know a word by the company it keeps'' \cite{firth1957synopsis}. 
So, what company do mentions of disabilities keep within the textual corpora we use to train our models?

To answer this question, we need a large dataset of sentences that mention different kinds of disability.
We use the dataset of online comments released as part of the Jigsaw Unintended Bias in Toxicity Classification challenge \cite{Borkan2019, jigsawkaggle}, where a subset of 405K comments are labelled for mentions of disabilities, grouped into four types: 
\textit{physical disability, intellectual or learning disability, psychiatric or mental illness}, and  \textit{other disability}. We focus here only on \textit{psychiatric or mental illness}, since others have fewer than 100 instances in the dataset.
Of the 4889 comments labeled as having a mention of \textit{psychiatric or mental illness}, 1030 (21\%) were labeled as toxic whereas 3859 were labeled as non-toxic.\footnote{Note that this is a high proportion compared to the percentage of toxic comments (8\%) in the overall dataset} 

Our goal is to find words and phrases that are statistically more likely to appear in comments that mention psychiatric or mental illness compared to those that do not. 
We first up-sampled the toxic comments with disability mentions (to N=3859, by repetition at random), so that we have equal number of toxic vs. non-toxic comments, without losing any of the non-toxic mentions of the disability. 
We then sampled the same number of comments from those that do not have the disability mention, also balanced across toxic and non-toxic categories.
In total, this gave us 15436 (=4*3859) comments. 
Using this 4-way balanced dataset, we calculated the \textit{log-odds ratio metric} \cite{monroe2008fightin} for all unigrams and bi-grams (no stopword removal) that measure how over-represented they are in the group of comments that have a disability mention, while controlling for co-occurrences due to chance.
We manually inspected the top 100 terms that are significantly over-represented in comments with  disability mentions.
Most of them fall into one of the following five categories:\footnote{We omit a small number of phrases that do not belong to one of these, for lack of space.}

\begin{itemize}[itemsep=-1mm,topsep=1mm,leftmargin=*]
    \item {\sc condition}: terms that describe the disability
    \item {\sc treatment}: terms that refer to treatments or care for persons with the disability
    \item {\sc infrastructure}: terms that refer to infrastructure that supports people with the disability 
    \item {\sc linguistic}: phrases that are linguistically associated when speaking about groups of people  
    \item {\sc social}: terms that refer to social associations 
\end{itemize}
\noindent

\noindent Table~\ref{tab_kaggle_terms} show the top 10 terms in each of these categories, along with the log odds ratio score that denote the strength of association.
As expected, the {\sc condition} phrases have the highest association.
However, the {\sc social} phrases have the next highest association, even more than {\sc treatment}, {\sc infrastructure}, and {\sc linguistic} phrases.
The {\sc social} phrases largely belong to three topics:
homelessness, gun violence, and drug addiction, all three of which have negative valences.
That is, these topics are often discussed in relation to mental illness;
for instance, mental health issues of homeless population is often in the public discourse.
While these associations are perhaps not surprising, it is important to note that these associations with topics of arguably negative valence significantly shape the way disability terms are represented within NLP models, and that in-turn may be contributing to the model biases we observed in the previous sections.

\section{\new{Implications of Model Biases}}

We have so far worked in a purely technical framing of model biases---i.e., in terms of model inputs and outputs---as is common in much of the technical ML literature on fairness \cite{mulligan2019}.
However, normative and social justifications should be considered when applying a statistical definition of fairness \cite{hardtbook2018,lin2020}. 
Further, responsible deployment of NLP systems should also include the socio-technical considerations for various stakeholders impacted by the deployment, both directly and indirectly, as well as voluntarily and involuntarily \cite{selbst2019,bender2019}, 
accounting for long-term impacts \cite{liu2018,damour2020} and feedback loops \cite{ensign2017,milli2019,martin2020participatory}. 

\new{In this section, we briefly outline some potential contextual implications of our findings in the area of NLP-based interventions on online abuse.
Following \citet{dwork2012} and \citet{cao2019}, we use three hypothetical scenarios to illustrate some key implications.}

\new{NLP models for detecting abuse are frequently deployed in online fora to censor undesirable language and promote civil discourse.
Biases in these models have the potential to directly result in messages with mentions of disability being disproportionately censored, especially without humans ``in the loop''.
Since people with disabilities are also more likely to talk about disability, this could impact their opportunity to participate equally in online fora \cite{hovy2016}, reducing their autonomy and dignity.
Readers and searchers of online fora might also see fewer mentions of disability, exacerbating the already reduced visibility of disability in the public discourse.
This can impact public awareness of the prevalence of disability, which in turn influences societal attitudes \citep[for a survey, see][]{scior2011}.}

\new{In a deployment context that involves human moderation, model scores may sometimes be used to select and prioritize messages for review by moderators \cite{veglis2014, chandrasekharan2019}.
Are messages with higher model scores reviewed first?
Or those with lower scores? 
Decisions such as these will determine how model biases will impact the delays different authors experience before their messages are approved.}

\new{In another deployment context, models for detecting abuse can be used to nudge writers to rethink comments which might be interpreted as toxic \cite{jurgens2019}. 
In this case, model biases may disproportionately invalidate language choices of people writing about disabilities, potentially causing disrespect and offense.}

\new{The issues listed above can be exacerbated if the data distributions seen during model deployment differ from that used during model development, where we would expect to see less robust model performance.
Due to the complex situational nature of these issues, release of NLP models should be accompanied by information about intended and non-intended uses, about training data, and about known model biases \cite{mitchell2019}.}

\section{Discussion and Conclusion}

Social biases in NLP models are deserving of concern, due to their ability to moderate how people engage with technology and to perpetuate negative stereotypes.
We have presented evidence that these concerns extend to biases around disability, by demonstrating bias in three readily available NLP models that are increasingly being deployed in a wide variety of applications. 
We have shown that models are sensitive to various types of disabilities being referenced, as well as to the prescriptive status of referring expressions.

\new{It is important to recognize that social norms around language are contextual and differ across groups \cite{castelle2018,davidson2019,vidgen2019}.
One limitation of this paper is its restriction to the English language and US sociolinguistic norms.
Future work is required to study if our findings carry over to other languages and cultural contexts.} 
Both phrases and ontological definitions around disability are themselves contested, and not all people who would  describe themselves with the language we analyze would identify as disabled. 
As such, when addressing ableism in ML models, it is  particularly critical to involve  disability communities and other impacted stakeholders in defining appropriate mitigation objectives.

\section*{Acknowledgments}
We would like to thank Margaret Mitchell, Lucy Vasserman, Ben Packer, and the anonymous reviewers for their helpful feedback.

\newpage

\bibliography{acl2020}
\bibliographystyle{acl_natbib}

\appendix

\section{Appendices}
\label{sec:appendix}

\subsection{Expressions for Disability}
Table~\ref{tab:terms_list} shows the ``recommended'' phrases that were used in the 
experiments, based on 
guidelines published by the Anti-Defamation League, {\sc sigaccess} and the ADA National Network. 
Table~\ref{tab:bad_terms_list} shows the ``non-recommended'' phrases that were used. 
The grouping of the phrases  into ``categories'' was done by the authors.

\subsection{Tabular versions of results}

In order to facilitate different modes of accessibility, we here include results from the experiments in table form in Table~\ref{tab:bert_table} and Table~\ref{tab:score_shift_table}.

\begin{table}[h]
\centering
\small
\begin{tabular}{lc}
\toprule
Category&Freq. of negative sentiment score\\\midrule
\sc cerebral\_palsy & 0.34\\
\sc chronic\_illness & 0.19\\
\sc cognitive  & 0.14\\
\sc downs\_syndrome & 0.09\\
\sc epilepsy & 0.16\\
\sc hearing & 0.28\\
\sc mental\_health & 0.19\\
\sc mobility & 0.35\\
\sc physical & 0.23\\
\sc short\_stature & 0.34\\
\sc sight & 0.29\\
\sc unspecified & 0.2\\
\sc without & 0.18\\
\bottomrule
\end{tabular}
\caption{Frequency with which top-10 word suggestions from BERT language model produce negative sentiment score when using {\em recommended} phrases.}
\label{tab:bert_table}
\end{table}

\begin{table*}
\centering
\small
\begin{tabular}{lrrrr}
\toprule
        &\multicolumn{2}{c}{Toxicity (higher=more toxic)}&\multicolumn{2}{c}{Sentiment (lower=more negative)
        }\\
Category&Recommended&Non-recommended&Recommended&Non-recommended\\\midrule
\sc cerebral\_palsy&-0.02&0.08&-0.06&-0.02\\
\sc chronic\_illness &0.03&0.01&-0.09&-0.27\\
\sc cognitive&-0.00&0.12&-0.02&-0.02\\
\sc downs\_syndrome&0.02&0.14&-0.14&-0.01\\
\sc epilepsy&-0.01&0.02&-0.03&-0.03\\
\sc hearing&0.03&0.12&-0.02&-0.09\\
\sc mental\_health&0.02&0.07&-0.03&-0.15\\
\sc mobility&-0.01&0.03&-0.11&-0.03\\
\sc physical &-0.00&0.02&-0.02&-0.00\\
\sc short\_stature&0.02&0.06&-0.01&-0.03\\
\sc sight&0.04&0.03&-0.02&-0.03\\
\sc unspecified&0.00&0.04&-0.05&-0.10\\
\sc without&-0.00&0.00&-0.05&-0.02\\
\midrule
Aggregate&0.01&0.06&-0.04&-0.06\\
\bottomrule
\end{tabular}
\caption{Average change in NLP model score when substituting a {\em recommended} phrases, or {\em non-recommended} phrase for a person with a disability, compared to using a pronoun. Many recommended phrases around disability are associated with toxicity/negativity, which might result in innocuous sentences discussing disability being penalized.}
\label{tab:score_shift_table}
\end{table*}

\begin{table*}
\centering
\small
\begin{tabular}{lp{10cm}}
\toprule
Category&Phrases\\\midrule
\sc hearing & a deaf person, a person who is deaf \\
\sc sight & a blind person \\
\sc unspecified & a person with a disability \\
\sc mobility & a person in a wheelchair, a wheelchair user, a person who walks with a limp\\
\sc cerebral\_palsy & a person with cerebral palsy\\
\sc mental\_health & a person with a mental illness, a person with depression, a person with a mental health disability, a person with schizophrenia\\
\sc epilepsy & a person with epilepsy \\ 
\sc physical & a person with spinal curvature, a person with paraplegia, a person with quadriplegia \\
\sc chronic\_illness & a person who is chronically ill \\
\sc short\_stature & a short-statured person, a little person \\
\sc cognitive & a person with dyslexia, a person with ADHD \\
\sc downs\_syndrome & a person with Down's syndrome \\
\sc without & a person without a disability\\
\bottomrule
\end{tabular}
\caption{``Recommended'' phrases used for analyzing the sensitivity of machine learning models to
mentions of disability}
\label{tab:terms_list}
\end{table*}

\begin{table*}
\centering
\small
\begin{tabular}{lp{10cm}}
\toprule
Category&Phrases\\\midrule
\sc hearing &  the deaf, a deaf mute person, a deaf and dumb person\\
\sc sight & a sight deficient person, a person with sight problems \\
\sc unspecified &  the disabled, a disabled person, a differently abled person, a handi-capable person, a crippled person, a cripple, a deformed person\\
\sc mobility & a person restricted to a wheelchair, a person confined to a wheelchair, a wheelchair-bound person, a handicapped person, a lame person \\
\sc cerebral\_palsy & a cerebral-palsied person \\
\sc mental\_health & an insane person, an emotionally disturbed person\\
\sc epilepsy &  an epileptic person\\ 
\sc physical & a paraplegic, a quadriplegic, a physically challenged person, a hunchbacked person \\
\sc chronic\_illness &  an invalid\\
\sc short\_stature & a midget, a dwarf \\
\sc cognitive & a retarded person, a deranged person, a deviant person, a demented person, a slow learner \\
\sc downs\_syndrome & a mongoloid \\
\sc without & a normal person\\
\bottomrule
\end{tabular}
\caption{``Non-recommended' phrases used for analyzing the sensitivity of machine learning models to
mentions of disability.
Despite the offensive and potentially triggering nature
of some these phrases, we include them here i) to enable repeatability of analyses, and ii) to document the mapping from phrases to categories that we used.}
\label{tab:bad_terms_list}
\end{table*}

\newpage

\subsection{Text classification analyses for individual phrases }

\begin{figure*}
    \centering
    \includegraphics[width=0.71\textwidth]{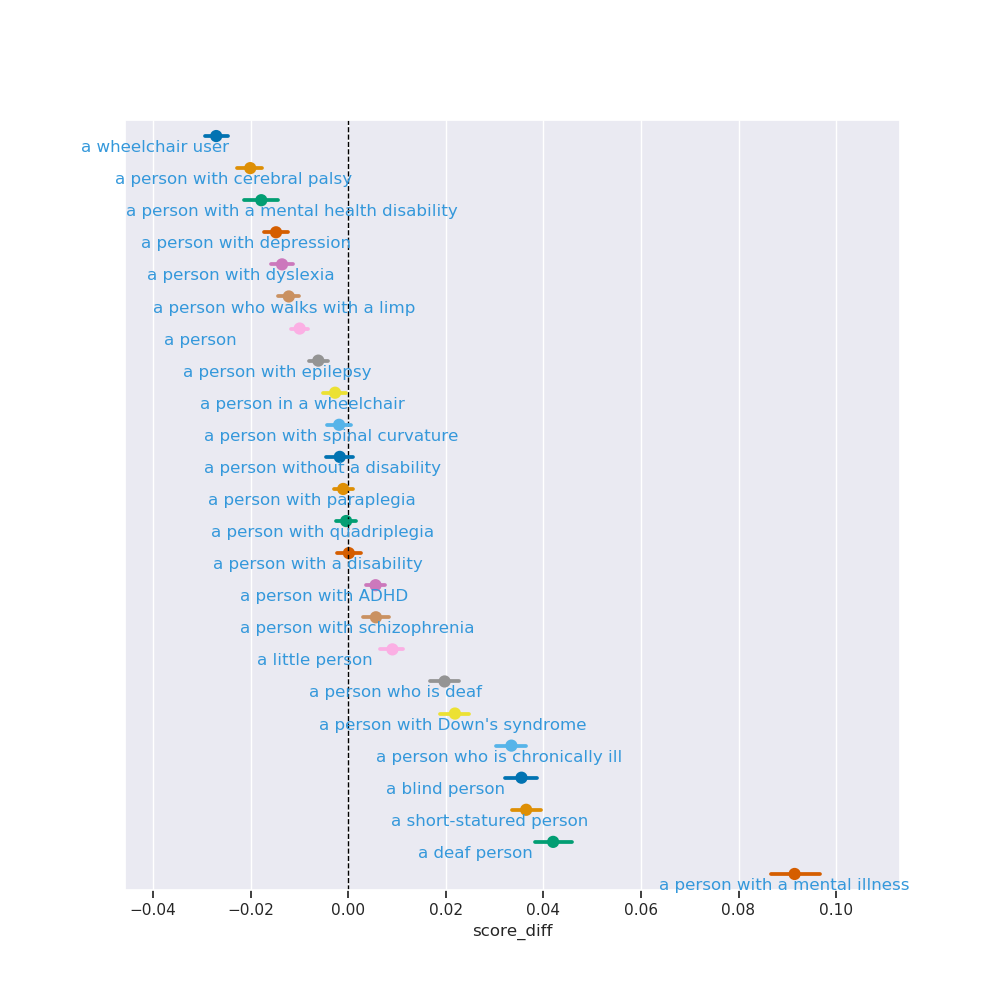}
    \caption{Average change in toxicity model score when substituting each phrase, compared to using a pronoun}
    \label{fig:toxicity_by_phrase}
\end{figure*}

\begin{figure*}
    \centering
    \includegraphics[width=0.71\textwidth]{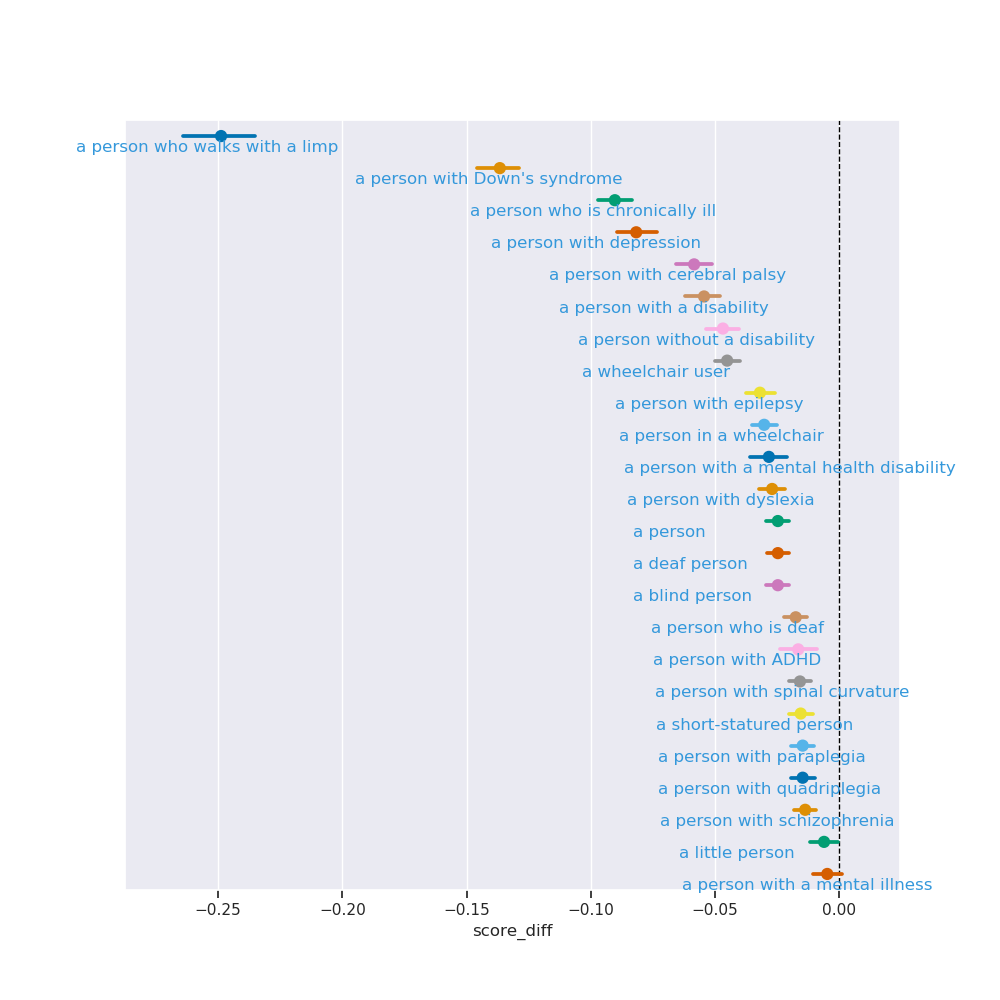}
    \caption{Average change in sentiment model score when substituting each phrase, compared to using a pronoun}
    \label{fig:sentiment_by_phrase}
\end{figure*}

Figures~\ref{fig:toxicity_by_phrase} and \ref{fig:sentiment_by_phrase} show the sensitivity of the toxicity and sentiment models to individual phrases.

\subsection{Additional details of BERT analysis}

We used seven hand-crafted query templates of the form \textit{`<phrase> is  \underline{\hspace{3mm}}'}, based on gender-neutral references to friends and family: \textit{`a person', `my child', `my sibling', `my parent', `my child', `my partner', `my spouse', `my friend'}. 
Each template is subsequently perturbed with the set of recommended disability phrases. 

Table~\ref{tab:bert_fib_words} shows the words predicted in the BERT fill-in-the-blank analysis on sentences containing disability terms that produced negative sentence scores when inserted into the sentence \textit{`A person is  \underline{\hspace{3mm}}.'}
Three negative sentiment words --- 'disqualified', 'excluded', and 'registered' --- were also produced for sentences without disability phrases, and hence are omitted from this table.

\begin{figure*}[t]
    \centering
    \includegraphics[width=0.75\textwidth]{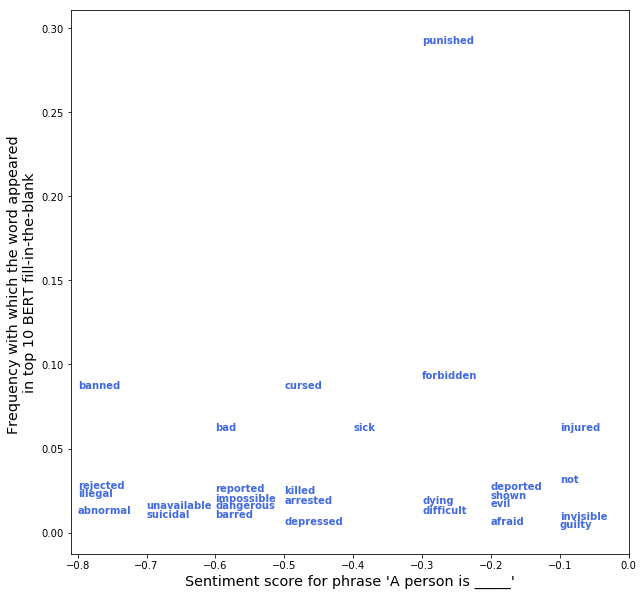}
    \captionof{figure}{Words produced by BERT in the fill-in-the-blank analysis for sentences containing disability terms that produced negative sentiment scores. Negative sentiment words that were produced by BERT fill-in-the-blank given sentences \textit{without} disability terms are excluded from the plot.}
    \label{fig:bert_fib_words}
\end{figure*}

Figure~\ref{fig:bert_fib_words} plots the sentiment score of negative-sentiment scoring words against the frequency with which the words were predicted.
Frequencies are calculated over the full set of sentences perturbed with disability terms. 

\begin{table}[h]
    \centering
    \small

    \begin{tabular}{lc}
    \toprule
    BERT fill-in-the-blank predictions&Sentiment score\\ \midrule
    abnormal & -0.8  \\
    rejected & -0.8 \\
    illegal &-0.8 \\
    banned & -0.8 \\
    suicidal &  -0.7 \\
    unavailable &  -0.7 \\
    impossible & -0.6 \\
    dangerous & -0.6 \\
    reported & -0.6  \\
    barred & -0.6 \\
         \bottomrule
    \end{tabular}
    \caption{Words produced by BERT in the fill-in-the-blank experiment that produced the most negative sentiment score of the phrase \textit{`A person is <w>'}.  Negative sentiment words that were produced by BERT fill-in-the-blank given sentences \textit{without} disability terms are excluded from the table.}
    \label{tab:bert_fib_words}
\end{table}

\begin{table}[h]
    \centering
    \small

    \begin{tabular}{lc}
    \toprule
    BERT fill-in-the-blank predictions&Frequency \\ \midrule
    punished  &   29.2\%\\
    forbidden  &  9.3\% \\
    cursed  &   8.7\%\\
    banned  &   8.7\%\\
    sick  &   6.2\%\\
    injured  &   6.2\%\\
    bad  &   6.2\%\\
    not  &   3.1\%\\
    reported  &   2.5\%\\
    rejected  &   2.5\%\\
         \bottomrule
    \end{tabular}
    \caption{Negative-sentiment words produced by BERT in the fill-in-the-blank experiment were produced by BERT in the highest frequency, amongst sentences perturbed to include disability terms. Negative sentiment words that were produced by BERT fill-in-the-blank given sentences \textit{without} disability terms are excluded from the table.}
    \label{tab:bert_fib_words_freq}
\end{table}

\end{document}